\documentclass[final,5p,times,twocolumn]{elsarticle}

\usepackage{graphics}
\usepackage{stackengine}
\usepackage{multirow}
\usepackage[usenames]{color}
\usepackage[table]{xcolor}
\usepackage{makecell}
\usepackage[strings]{underscore}

\usepackage{tabularx}
\newcolumntype{Y}{>{\centering\arraybackslash}X}
\usepackage{hhline}
\usepackage[normalem]{ulem}

\usepackage{float}
\usepackage{amssymb}
\usepackage{amsmath}
\journal{Journal of Petroleum Science \& Engineering} 

\begin{document}

\begin{frontmatter}

\title{Data-driven model for hydraulic fracturing design optimization. Part II: Inverse problem}


\author[SKT]{V.M.~Duplyakov}
\author[SKT]{A.D.~Morozov}
\author[SKT]{D.O.~Popkov}
\author[GPN]{E.V.~Shel}
\author[SKT]{A.L.~Vainshtein}
\author[SKT]{E.V.~Burnaev}
\author[SKT]{A.A.~Osiptsov}
\ead{a.osiptsov@skoltech.ru}
\author[GPN]{G.V.~Paderin}

\address[SKT]{Skolkovo Institute of Science and Technology (Skoltech), 3 Nobel Street, 143026, Moscow, Russian Federation}
\address[GPN]{Gazpromneft Science \& Technology Center, 75-79 liter D Moika River emb., St Petersburg, 190000, Russian Federation}

\begin{abstract}
We describe a stacked model for predicting the cumulative fluid production for an oil well with a multistage-fracture completion based on a combination of Ridge Regression and CatBoost algorithms. The model is developed based on an extended digital field data base of reservoir, well and fracturing design parameters. The database now includes more than 5000 wells from 23 oilfields of Western Siberia (Russia), with 6687 fracturing operations in total. Starting with 387 parameters characterizing each well, including construction, reservoir properties, fracturing design features and production, we end up with 38 key parameters used as input features for each well in the model training process. The model demonstrates physically explainable dependencies plots of the target on the design parameters (number of stages, proppant mass, average and final proppant concentrations and fluid rate). We developed a set of methods including those based on the use of Euclidean distance and clustering techniques to perform similar (offset) wells search, which is useful for a field engineer to analyze earlier fracturing treatments on similar wells. These approaches are also adapted for obtaining the optimization parameters boundaries for the particular pilot well, as part of the field testing campaign of the methodology. An inverse problem (selecting an optimum set of fracturing design parameters to maximize production) is formulated as optimizing a high dimensional black box approximation function constrained by boundaries and solved with four different optimization methods: surrogate-based optimization, sequential least squares programming, particle swarm optimization and differential evolution. A recommendation system containing all the above methods is designed to advise a production stimulation engineer on an optimized fracturing design.
\end{abstract}

\begin{keyword}
hydraulic fracturing \sep machine learning \sep predictive modelling \sep design optimization \sep gradient-free optimization \sep probability of improvement \sep surrogate optimization \sep multistage fracturing
\end{keyword}

\end{frontmatter}

\section{Introduction and problem formulation}
\label{sec1}

In this paper, we continue the work on the development of a workflow on hydraulic fracturing (HF) design optimization with machine learning on field data, which was initiated in~\cite{MLforHF2020}. In Part I of this project, we presented the detailed methodology for development of a digital database on reservoir and well parameters, design parameters of multistage fracturing treatments and production data. We also discussed in detail the forward problem of production forecast based on reservoir, well and fracturing design data. Here we continue the effort and now move on to the development of the optimization workflow (inverse problem).

Before we move to our results, let us review the recent developments in the area of machine learning-assisted optimization of hydraulic fracturing designs. In~\cite{MLforHF2020}, we made a thorough review of the relevant literature published on this subject, so the reader is referred to~\cite{MLforHF2020} for the state of the art, whereas here we will only mention new papers appeared after Part I of our study has been published.

The paper~\cite{RN2019refrac} uses candidate selection for refracturing operations based on data analytics. Gradient boosting is used for production forecast, and the feature importance analysis indicates the current production (before refracturing) to be the most important parameter. We would like to emphasize that the problem of candidate selection for refracturing treatments and the problem of primary fracturing design optimization are the two very different problems by formulation: the prior production is known when it comes to refracturing, which makes the task of predicting subsequent production much easier.

In~\cite{xue2019shales}, the multiobjective random forest method is proposed to predict the dynamic production data for a shale gas well, with geological and hydraulic fracturing properties used as input features. The performance of multi-objective random forest (MORF) and multi-output regression chain (MORC) methods are compared. A dynamic regression model is constructed in~\cite{yandex2020oil} using a machine learning method referred to as the sliding window regression. Features of the model are divided into four groups: integral, local, pressure and autoregressive. The idea of sliding window is used to obtain a model with stable coefficient dynamics and ability for long-term forecast. 

In~\cite{yao2021optimization}, variable-length particle-swarm optimization (Modified Variable-length PSO, MVPSO) was proposed to automatically select the optimal fracturing parameters: the number of fractures as well as the corresponding fracture properties. Then, MVPSO was verified and compared with VPSO by several benchmarks. In addition, a gas/water two-phase model considering gas-adsorption and Knudsen-diffusion effects was used to describe the shale-gas flow in matrix and fracture domains.

About this work: the paper is organized as follows. The status of the development of a digital database is summarized in Sec.~\ref{sec2}. We describe  the model for production forecast (forward problem)  in Sec.~\ref{sec3}. A study on offset wells selection for finding optimization intervals for fracturing design parameters is presented in Sec.~\ref{sec4}. Inverse problem of finding the optimized fracturing design parameters is discussed in Sec.~\ref{sec5}. Overall discussion and analysis of the results are given in Sec.~\ref{sec6}. The paper ends up with summary and conclusions in Sec.~\ref{sec7}.

\section{Data}
\label{sec2}
\subsection{Database overview}
\label{sec2.1}
In this work, we use the version 7.9 of the database. Compared to the database version 5.5 used in our previous publication~\cite{MLforHF2020}, the overall architecture and the sources of data used to compile the database for production forecast have not been changed. The database still contains information on more than six thousands miscellaneous fracturing operations (single-stage and multi-stage, primary and re-fracturing) or about 17 000 injections on 5 425 wells from 23 different fields in Western Siberia for the period 2013-2019, including well profile, fracturing design, geological and production reports.  Also, the centralized approach to data organization is remained unchanged.

Our work on improvements of the database have been continued and the quality of the preprocessed data was increased so far through consultations with engineers ingrained in the raw data collection process.  

\begin{figure}[h!]
\includegraphics[width=8cm]{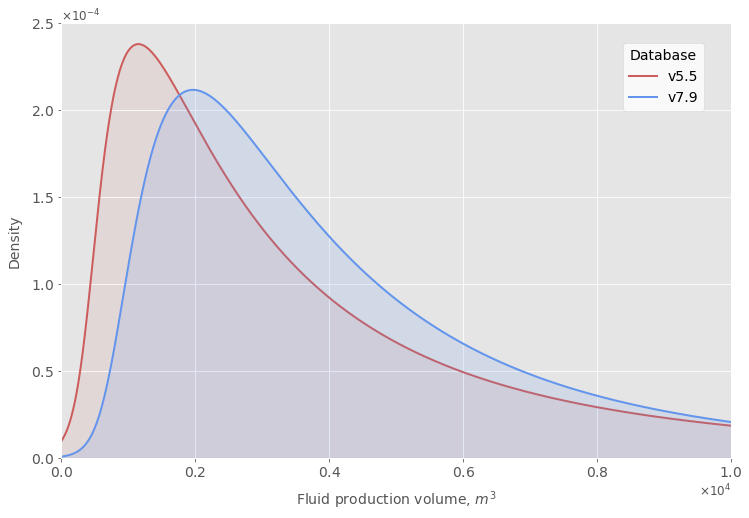}
\caption{Target probability distributions for database v.~5.5 used in~\cite{MLforHF2020} (red) \& the current database v.~7.9 (blue).}
\label{Target_distribution}
\end{figure}

Current version of the database 7.9 contains the following changes:
\begin{itemize}
    \item Target preprocessing is updated: monthly data consolidated in 3-, 6- and 12-months slices are summed also for all active reservoirs on the well and not only the stimulated ones. The shape of the target distribution, predictably, has become smoother by moving the peak from almost 1,000 to near 2,000 cubic metres of fluid, as shown in Fig.~\ref{Target_distribution};

%Before, the features were calculated in proportion to layer net pay;
    \item There was an issue with the lack of standardization in reservoir's and well's namings among the data sources. Similarity between labels is now achieved through morphological analysis;
    \item Added new features: proppant properties (grain size and density of prevailing type by proppant manufacturer);
    \item Added facies for the dominant field. The feature is described by five zones from lithological maps: slope zone, proximal fan, shallow area, alluvial fan and depocenters.
% зона склона, проксизона, мелководье, конусы, депоцентры
\end{itemize}

%The authors also draw attention that previously described version of the database (v5.5) is deprecated and no longer in use for now.

\subsection{Target variable}

Selection of the right target variable is crucial for the success of the entire optimization workflow, so we think it deserves a separate discussion in this subsection. We analyzed the data made available to us and also discussed in detail the current strategies of fracturing design development among production stimulation engineers and come to a conclusion that fracturing operations are optimized in the field based on the metrics of maximizing reservoir contact: the larger the fracture the better. Hence, bigger fractures yield larger cumulative volume of recovered total fluid (both oil and reservoir water), and there is a strong  correlation. Rarely this consideration takes into account the fact the excessively large fractures may breakthrough into upper or lower layers, thereby causing additional production of water. Thus, a model targeted to predict the cumulative volume of total produced fluid was trained on existing data with slightly higher accuracy (as there is a direct correlation between the input parameters characterizing the fracturing design and the total cumulative produced fluid), as compared to a model targeted on cumulative oil only, where the prediction accuracy was lower, as the oil production is not directly correlated with the design parameters governing the fracture dimensions. The present realization of the model does not take into account (yet) the presence of water-bearing layers, so the model better predicts the total volume of produced fluid. During the pilot works, we deal with the particular oilfield where there are no bottom waters nearby target formation, so the targets of cumulative total fluid and cumulative oil are practically equivalent.  

At the same time, we realize that generalization of the present workflow to oilfields with the presence of aqueous layers will require modifications: we will need to add some features characterizing the presence of upper/lower water bearing formations to be able to predict the production of oil and water separately, and also the target variable should be composed of two components: maximum total fluid and maximum pure oil (or minimum water cut, which is equivalent).

We use the 3-month cumulative production data to utilize all the data available, including the most recently fractured wells, where the production history is short.  

%The later fracturing was carried out, the less production history we know. So, the shorter the considered period - the more HF experience is expressed in the dataset. Therefore, 3-month (90 days) production data slices were chosen as representative sets.
%Which is why the most recent operations were not taken into account: there are known only 1-2 months production data. 

An interpolation of the monthly data was used in order to come up with 90-day production period. This approach is suitable, for instance, if a well worked for 20 days in the first month after fracturing treatment and 30 days in the following months, the desired value is in 80-110 days or 3-4 months range and can be found through the interpolation.

\subsection{Feature selection}

In the first part of this study~\cite{MLforHF2020}, we made efforts to utilize the entire database with primary fracturing operations, as well as refracturing to solve the forward problem of predicting the production rate without any preliminary separation of the dataset.  

Currently, the database is divided into two parts: primary and repeated stimulations. We fitted models for prediction of the fluid production for both these types of operations. Further, during the field tests, we focused only on primary stimulation treatments optimization.

%Nevertheless, we would like to emphasize that the resources of the initial database allow to build two different models on two subsets: production prediction of primary fracturing operations and repeated stimulation treatments (refracturing). 

To construct the models and increase interpretability of the problem we use feature importance analysis. It can be done with or without the involvement of an approximation model.  

\begin{figure}[H]
\includegraphics[width=8cm]{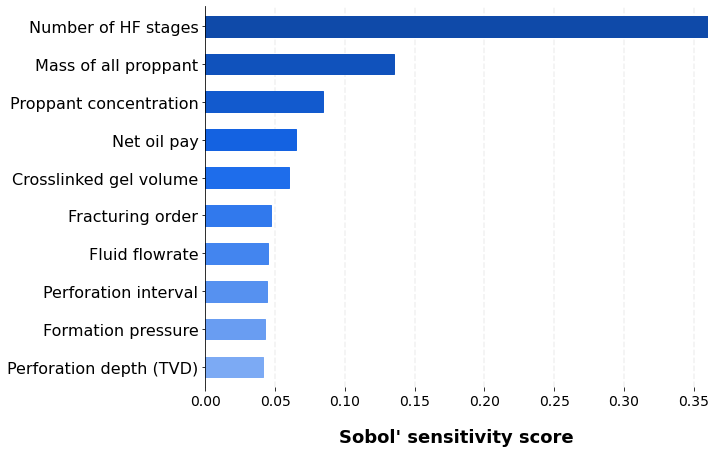}
\caption{Sobol sensitivity for the entire database.}
\label{sobol}
\end{figure}

One method is a sensitivity analysis via Sobol indices~\cite{sobolPaper}, which can be calculated without constructing any approximation model and decompose the variance of the target variable into parts attributed to input features. The most important features can be seen in Fig.~\ref{sobol}. 

On the contrary, SHAP method~\cite{SHAP} is based on an approximation model, utilizes the concept of Shapley values~\cite{shapley1953value} and measures features importances in terms of predictive power of each feature. The use of this method more accurately shows the true importance of the features~\cite{song2016shapley}. The SHAP values can be calculated for tree-based models (which we use).

\begin{figure}[h!]
\includegraphics[width=8cm]{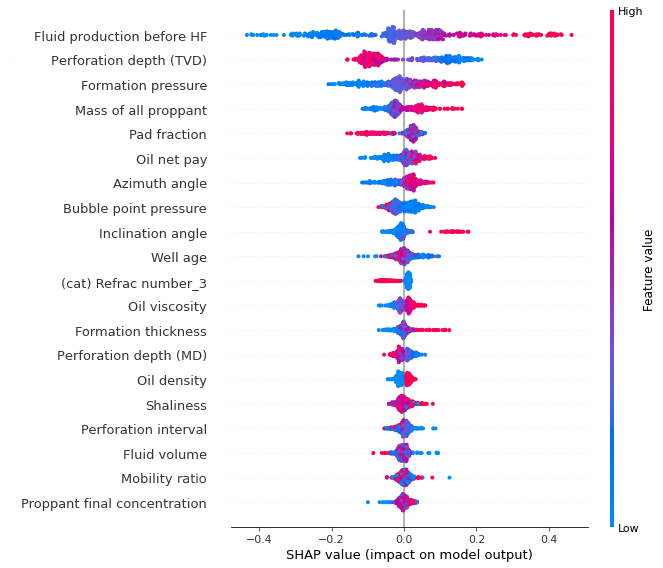}
\caption{SHAP feature importance: refracturing operations.}
\label{Feature importance_refrac}
\end{figure}

\begin{figure}[h!]
\includegraphics[width=8cm]{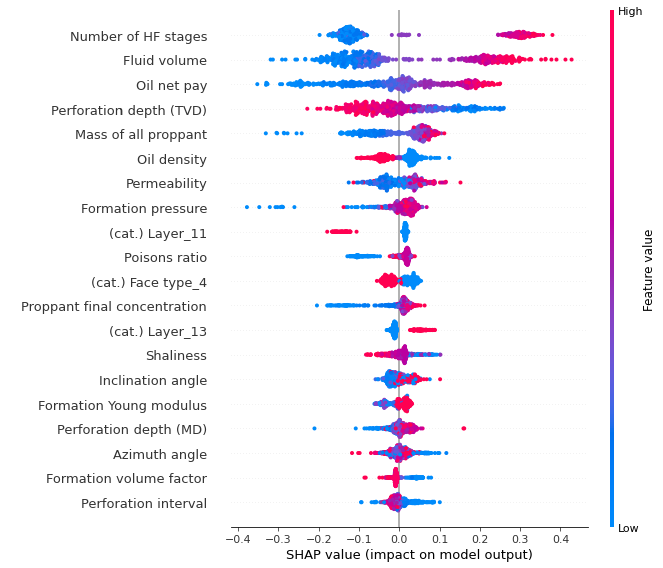}
\caption{Feature importance: primary fracturing on new wells.}
\label{Feature importance_new}
\end{figure}

Indeed, the difference between the two models for primary stimulation vs refracturing can be seen through the SHAP feature importance analysis (Figs.~\ref{Feature importance_refrac},~\ref{Feature importance_new}). (In case of  categorical features we use one-hot encoding. The corresponding features are prefixed with ``cat.''.) Particularly, for refracturing operations the key feature having major impact on the target is the level of production before the refracturing treatment (which was not captured in previous analysis), which is absent in case of primary operations. Having data on production prior to refracturing makes the production forecast problem easier to solve, compared to the case of production forecast after primary fracturing operations. This has also been noted in other studies~\cite{erofeev}.

Based on the analysis, we may conclude that the features describing proppant properties (introduced in Sec.~\ref{sec2.1}) are not important for predicting the fluid production.  

Another way of analyzing parameters is the feature elimination procedure which implies a reduction of feature space which in its turn may improve the performance of the approximation model. Feature elimination can be achieved by applying a pair-wise correlation of parameters via the Spearman correlation and by the Recursive Feature Elimination (RFE) method which involves the use of the approximation model. Speaking of the former, the reason of choosing the Spearman correlation instead of the Pearson correlation is because most of the features correlations are non-linear. So, by estimating correlations between features we can remove perfectly correlated features and features with zero variance. Regarding the second method, RFE is a procedure for backward selection of features which works as follows. First, a model is built on an the entire set of input features, and features' importance is calculated. Then, the least important features are removed. After that, the model is rebuilt on the reduced set of features and we repeat the process.

Eventually, we use the following procedure for feature elimination:
\begin{enumerate}
    \item Select 3-month slices; (6- and 12-months  production \& geological and technical data are removed due to the lack of data from the latest treatments);
    \item Remove parameters which are not relevant for the considered target variable, or for which more than 80\% of the observations are missing;
%As the result, we removed \textcolor{red}{how many} from \textcolor{red}{how many} features;
    \item Estimate Spearman correlations to identify perfectly correlated features; remove features with almost zero variance;
    \item Apply RFE to select features, which are the most important for the target prediction.
\end{enumerate}

As the result,  we reduced our initial set of features from 387 to 38 features.

\section{Forward problem of oil production forecast after fracturing.}
\label{sec3}

\subsection{Problem formulation}
\label{sec3.1}
In this section, we describe the methods developed for prediction of the cumulative fluid production. Previously ~\cite{MLforHF2020}, we created prediction model based on gradient boosting algorithm which was based on various features including reservoir parameters, well construction and HF design data (which includes many features like pumping pressure, fluid efficiency, fracture geometry, etc). Those features were useful for predicting the efficiency of HF. However, in this paper, we focus on choosing the proper HF design only for the new wells. We need to know the optimal design before the treatment, therefore we cannot use all the design features, as many of them are known only after the fracturing. Chosen features for the optimization are:
\begin{itemize}
    \item Number of stages,
    \item Pad share,
    \item Fracturing fluid volume (or average proppant concentration),
    \item Proppant mass,
    \item Fluid rate,
    \item Final proppant concentration (proppant concentration at the last stage of pumping).
\end{itemize}
These parameters are essential for planning the HF treatment. Through those features one can estimate a pumping schedule for a particular well.  

One may notice the absence of one of the most important parameters, which determines the fracturing fluid viscosity --- polymer concentration (in water-based polymer gels). This is due to the very low variability of this parameter in the initial database.

% \subsection{Uncertainty quantification}
% % catboost license посмотреть как цитировать
% Gradient boosting regression models return a point prediction for a given input vector of features. A recent update of the Catboost algorithm introduced a new capability of quantifying uncertainty of the prediction~\cite{catboostunc}. Moreover, uncertainty obtained from Catboost can be decomposed into data and knowledge uncertainties as opposed to NGBoost~\cite{ngboost} where it only estimates the data uncertainty. Nevertheless, such functionality for quantifying prediction uncertainty is very useful for our problem since we are dealing with the data with relatively low fidelity  while literally having no rights for errors in our predictions. Otherwise, wrong design recommendations may lead to costly financial losses.  

\subsection{Stacking}
The most successful algorithms in terms of error metrics were decision-tree based ones for the problems with heterogeneous set of features. For our prediction task, the CatBoost outperformed all others.  

\begin{figure*}[h!]
\centering
\includegraphics[width=16cm]{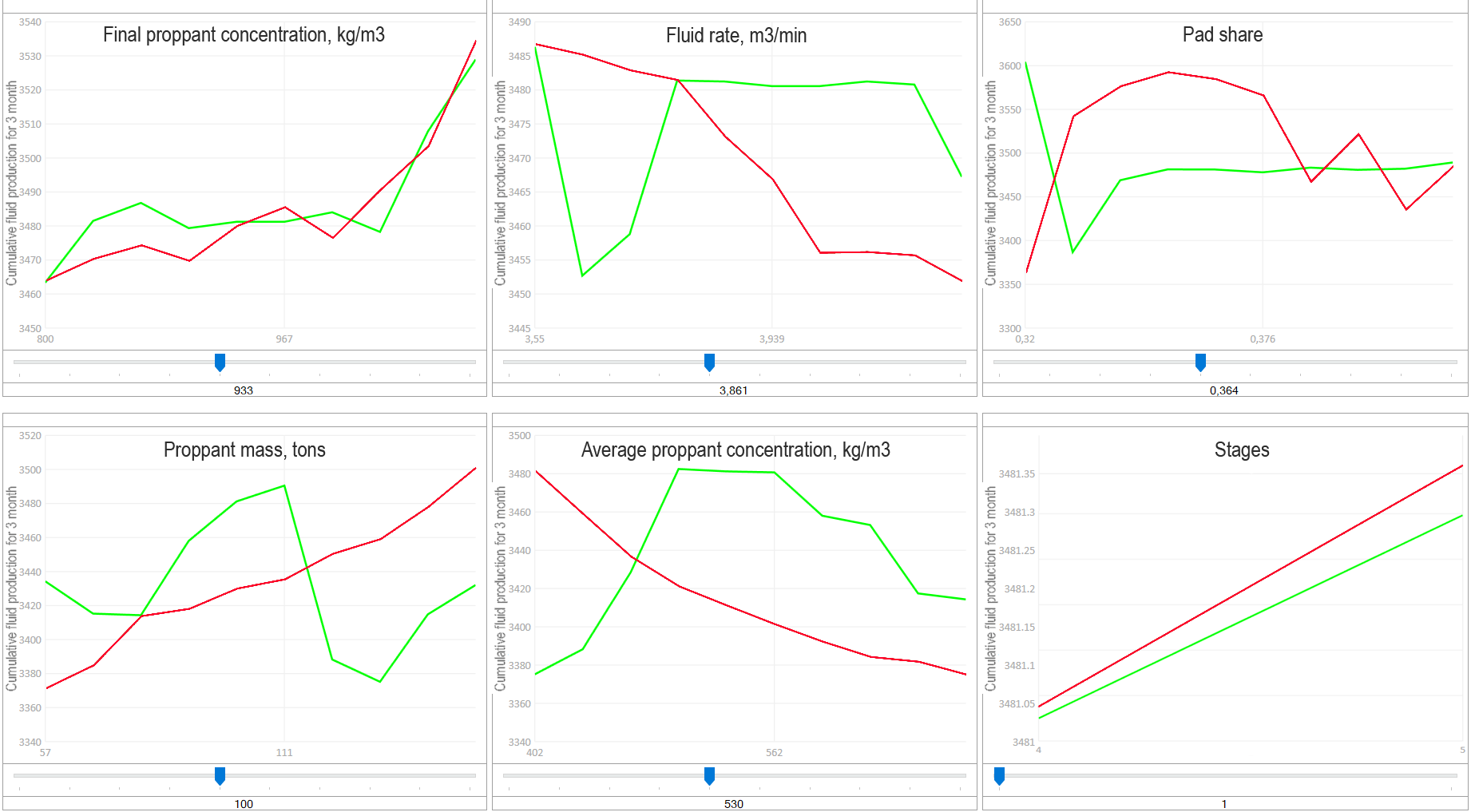}
\caption{{CatBoost (green) vs Stacked (Ridge+CatBoost) dependences}}
\label{stackedVScatb}
\end{figure*}

Besides, the model predictions should be physically explainable and robust. To check these properties we plotted dependencies of the target on the input design parameters. We expect the plots to demonstrate smooth and robust  behaviour. One can see in Figure~\ref{stackedVScatb} (green line) that CatBoost-based plots are not smooth, as there is not so much initial data and decision-tree based algorithms create piece-wise constant approximations.  

\begin{figure}[H]
\centering
\includegraphics[width=8cm]{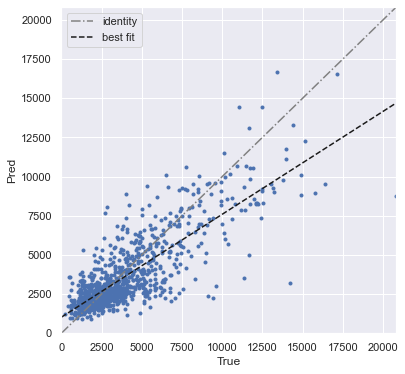}
\caption{Scatter plot for the stacked model}
\label{regr_plot}
\end{figure}

To correct this  behaviour of the model we used a stacking approach. In the first step, we trained a ridge regression model, then we subtracted predictions of the model from the true target values and then fitted a CatBoost regression model on the residuals. Hyperparameters tuning was performed using 5-folds cross validation. Overall model uses both ridge and CatBoost regression algorithms, summing the predictions. One can see the improved dependencies plots in Figure~\ref{stackedVScatb} (red line). We believe that these plots are more physically explainable. To characterize quality of the model we used scores presented in Table~\ref{tab:metrics}. These metrics were calculated on the hold-out set that is 30\% of the initial data. See also  the scatter plot in Fig.~\ref{regr_plot}.

\begin{table}[H]
\centering
\begin{tabular}{|l|c|}
\hline
{\cellcolor[HTML]{EFEFEF}Quality score} & {\cellcolor[HTML]{EFEFEF}Value}  \\ \hline\hline
R\textsuperscript{2}                                                   & 0.64    \\ \hline
MAE                                                  & 1131    \\ \hline
MedianAE                                             & 723.7  \\ \hline
MAPE                                                 & 36.08    \\ \hline
Weighted MAPE~\cite{wMAPE}                           & 29.07   \\ \hline
MSE                                                  & 2802631 \\ \hline
RMSE                                                 & 1674.11 \\ \hline
\end{tabular}
\caption{Quality scores for the stacked model}
\label{tab:metrics}
\end{table}

\section{Selection of optimization intervals}
\label{sec4}

After constructing the model, the inverse problem is formulated as finding a set of optimal fracturing design parameters to maximize our target. {Since the model is multi-dimensional, a valid selection of the design parameter values is only possible if they change within the relevant intervals during the optimization procedure. These intervals are caused by various constraints, arising in the field.}

{The additional restrictions on design parameter values can be divided into geological and technological constraints. Geological constraints include those related to the geological structure of the formation. For example, proximity of gas or water bearing formations leads to the necessity to limit fracture height growth, which automatically leads to limitation of the maximum volume of injected proppant and requires change of the perforation strategy. Geological constraints can also be related to waterflooding cases. For example, when an injection well, operating at bottomhole pressures higher than the formation breakdown pressure, is located relatively close to the production well and is in the direction of fracture propagation, it is necessary to limit the maximum fracture half-length on the production well.}

{Technological limitations include those related to the technical capabilities of the equipment and chemicals used. For instance, the maximum fracturing pressure (first of all, it is connected with the capabilities of the wellhead equipment and pumping units) can lead to the restriction on the maximum fracture width. Those limitations at a first approximation could be obtained by offset (similar) well search methods.}

{
In the field, a fracturing job is pumped with variable proppant concentration, so another limitation we add is a parameter characterising the rate of increase in the proppant concentration in the fracturing fluid from initial ($c_{start}$) to final ($c_{fin}$) concentration $[kg/m^3]$ with respect to increase to average $c_{avg}$. The parameter is denoted as $\epsilon$ and is defined as:
    \begin{equation}
        \epsilon = \frac{c_{fin} - c_{start}}{c_{avg} - c_{start}} - 1.
    \end{equation}
Boundaries for this parameter are the same for all pilot wells: from 0.5 to 1.5. This is not a parameter that needs to be optimized, we add these bounds as a constraint for optimization algorithms.
}

\subsection{Clustering for offset wells selection}
\label{boundaries}

%\subsubsection{Pilot cluster identification}
To find optimization intervals for fracturing design parameters we define a pilot cluster as a set of wells which are similar to the pilot one (the same field, layer, face and direction). We estimate design parameters limits as $5^{th}$ and $95^{th}$ percentiles of the values of the parameters for wells, belonging to the constructed cluster. Thus, we believe, the results of the optimization would be more robust, as we do not use extrapolated model's predictions.

\begin{figure}[H]
\centering
\includegraphics[width=8.8cm]{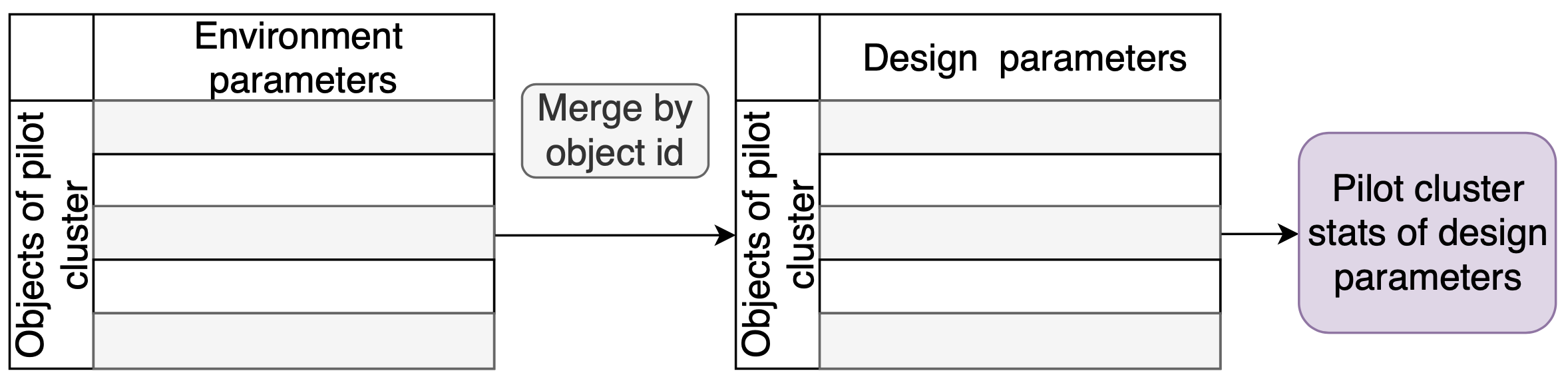}
\caption{Cluster procedure: steps 6 – 8}
\label{Cluster procedure stats}
\end{figure}

Another approach to obtain optimization boundaries for the specified pilot well is to utilize unsupervised ML methods like clustering and dimensionality reduction. The procedure is as follows (See also figs.~\ref{Cluster procedure}, \ref{Cluster procedure stats}):

\begin{figure}[H]
\centering
\includegraphics[width=8.8cm]{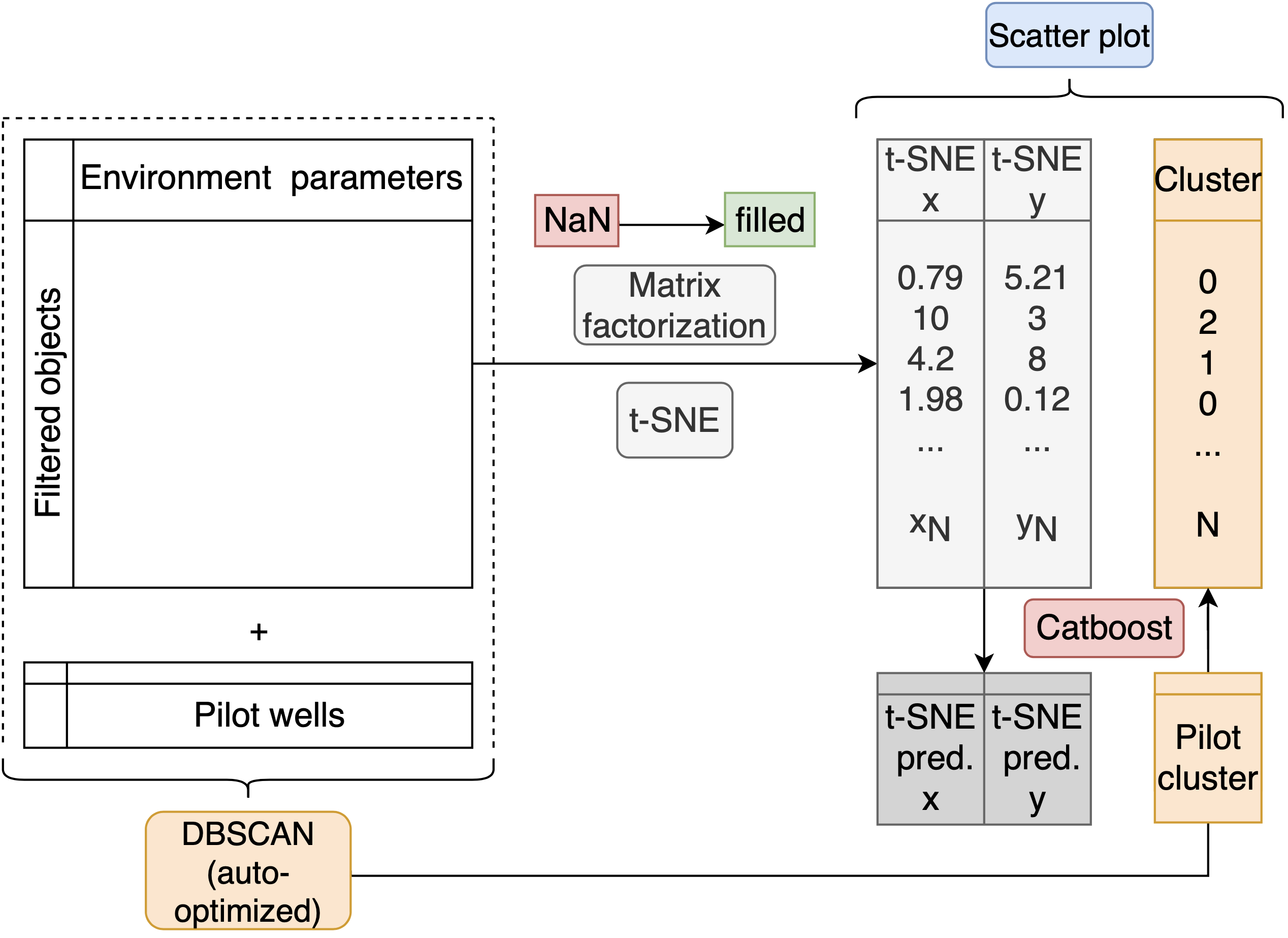}
\caption{Cluster procedure: steps 1 – 5, 9}
\label{Cluster procedure}
\end{figure}

\begin{figure}[h!]
\includegraphics[width=8cm]{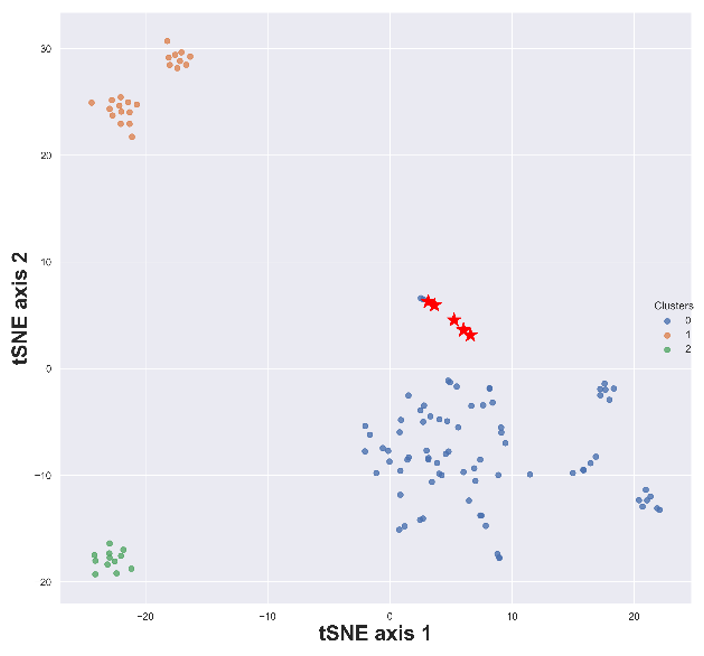}
\caption{t-SNE scatter plot for cluster visualization}
\label{t-SNE scatter}
\end{figure}

\begin{enumerate}
    \item Remove all features from the database except the environment parameters: PVT, geomechanics, well log interpretation, well id;
    \item Filter the obtained subset by the number of stages, layer id and face of the pilot well. This step is needed in order to collect wells technically similar to the pilot well;
    \item Apply a clustering algorithm to the environment parameters of the filtered data subset. Hyperparameters of the clustering method are optimized via a gradient-free optimization algorithm by maximizing the mean silhouette coefficient, calculated for a particular cluster as follows:
    \begin{equation}
        S_i=\frac{b_{i}-a_{i}}{max(a_i,b_i)},
    \end{equation}
    where $a_i$ is a mean intra-cluster distance, $b_i$ is a  mean nearest-cluster distance;
    \item Find a cluster to which the pilot well belongs to. Assert this cluster as a pilot well cluster;
    \item Using id-s of the objects from the pilot cluster, add the design parameters values to the objects' descriptions;
    \item Analyze the pilot cluster statistics: minimum, maximum and mean values of the design parameters. The minimum values and the maximum values serve as the optimization boundaries while the mean values of the design parameters are used as an initialization for the optimization problem.
\item Perform a visual analysis via the t-SNE algorithm based on the environment parameters. t-SNE can only be applied to data without missing values. Hence, the  missing values are imputed by the matrix factorization algorithm~\cite{MLforHF2020}. After that we obtain the t-SNE embedding for the selected subset of data. Here t-SNE embedding is a mapping of the multidimensional input features to the two-dimensional $x$ and $y$ coordinates for each observation of the data subset. These two-dimensional coordinates can be used for visualization of the considered data subset.
   \item After that, using the considered subset of the data and the obtained t-SNE coordinates  build an ML regression model to predict t-SNE $x$ and $y$ coordinates for any new object. The corresponding ML model takes as input environment parameters and returns as output t-SNE coordinates $x$ and $y$. 
   The reason of predicting the t-SNE coordinates is because any change in the data like appending a new pilot well to the data set will require running the  t-SNE algorithm from scratch;
    \item Predict the t-SNE coordinates $x$ and $y$ for the pilot well using the constructed regression model. This allows to visualize the position of a new well with respect to the selected data subset; %The reason of predicting the coordinates is because any change in data like appending a pilot well to the frame will yield a completely different transformation of the t-SNE algorithm as compared to the dataframe without a pilot well. That is why we would like to run t-SNE on the dataframe without appended pilot wells and then predict their t-SNE coordinates;
    \item Create a t-SNE scatter plot from the $x$ and $y$ t-SNE coordinates. Label the corresponding clusters of each object by some colors and label the pilot well by a star symbol. This procedure visually verifies how the data is clustered and how the pilot well is located with respect to the remaining data set.
\end{enumerate}
The example of this method implementation is shown in Fig.~\ref{t-SNE scatter}.

\subsection{Missing pilot parameters imputation}
In practice some of the parameters values of a well can be missing. 
Imputation is needed for wells parameters values because gradient-free optimization algorithms are based on constructing regression models and so they cannot work with missing values. For better results, we need to impute these values not just by filling them with corresponding mean values, but we should do it in a smarter way. We propose several strategies to do it. The matrix factorization method was described in the first part of our study~\cite{MLforHF2020}. The second method is to impute missing values by averaging parameters values of the top-\emph{N} similar wells. The method to find similar wells is described in Sec.~\ref{offsetwells}. Another way to make the imputation is to use the mean pilot cluster parameters values as described in step 6 of the algorithm in the previous subsection.

\subsection{Offset wells selection by Euclidean distance}
\label{offsetwells}

Offset wells are the wells, similar to the pilot one in terms of their geological surroundings. One may look for these wells both in terms of their geological and geographical (closest wells within certain radius from the pilot one) similarities. These analogue wells search is very useful for a petroleum engineer as it allows to analyse fracturing operations, conducted previously, and the design parameters values, check whether an operation was successful or not, etc. {We can also extract additional features from the neighbouring wells, which increase predictive power of the models~\cite{erofeev}. For example, in our work, we used features, such as average fluid production divided by distance from the pilot well, using wells within 1 km from the pilot one.}

\begin{figure}[h]
\centering
\includegraphics[width=16cm]{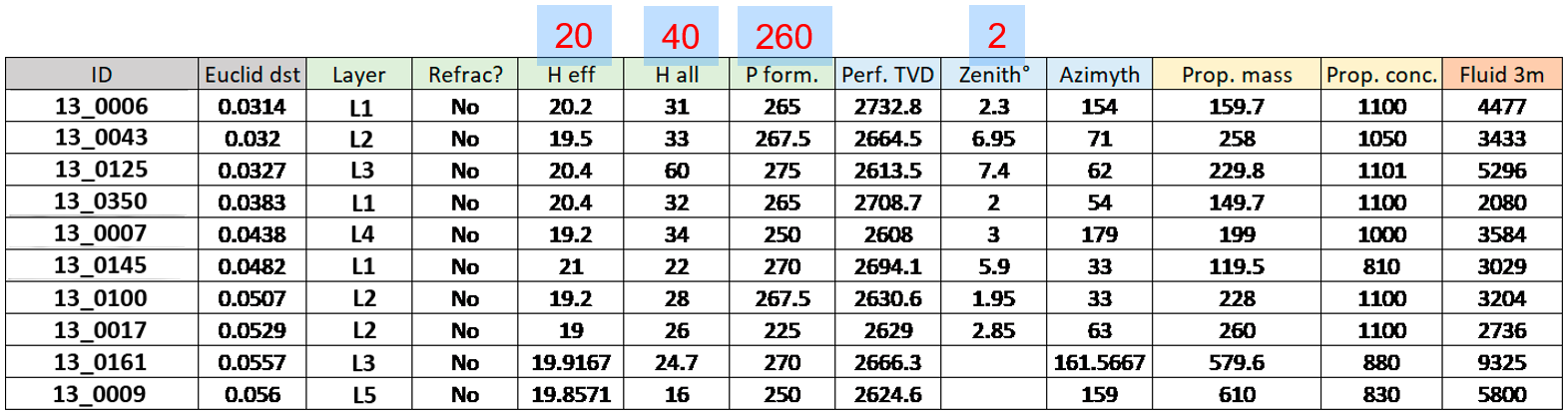}
\caption{Example of offset wells selection}
\label{simwellsearch}
\end{figure}

As a similarity metric we can use the Euclidean distance between the pilot well and the other wells. To calculate it, we firstly need to normalize values in the database. We perform a linear min-max normalization, where min and max values are the $1^{st}$  percentile and the $99^{th}$ percentile respectively. We use  percentiles to discard possible outliers in the initial data set. Then, the Euclidean distance between two vectors of parameters, characterizing wells, is calculated as
\begin{equation}
    d(p,q)={\sqrt{(p_{1}-q_{1})^{2}+\cdots +(p_{n}-q_{n})^{2}}},
\end{equation}
where $p_i$ and $q_i$ are the $i$-th parameter's values of the corresponding wells. The distances are calculated between the pilot well and all other wells, belonging to the same cluster (within the same field, layer and face). The results of such similar wells search are considered robust and sustainable by field geologists. One can see an example of a result of such search in Fig. \ref{simwellsearch}.

{
In this work we combine the clustering method for selecting offset wells with the euclidean distance search. The clustering is used for obtaining the set of similar wells, then we reduce the size of this set, leaving only top-N wells by the euclidean distance. This method allows us to look for optimal values of the design parameters in a certain vicinity where the prediction model works well. Also, in some cases the top-N similar wells can be used for imputing missing parameters for the pilot well.
}

% provide comparison

\section{Inverse problem of finding frac design parameters to maximize production}
\label{sec5}

\subsection{Problem formulation}
An inverse problem can be formulated as optimizing a high dimensional black box (BB) with respect to inputs constrained by boundaries. In our case BB is a function with unknown  expression or internal structure that, given a list of inputs, returns corresponding outputs. The high dimensionality of the input presents an exponential difficulty for problem modeling and optimization (so-called ``curse-of-dimensionality'')~\cite{bellman2015applied}. To optimize a high-dimensional computationally expensive black box (HEB) function, it is required to iteratively evaluate an objective function, which can be costly and so becomes unacceptable. In our case, the HEB function is represented by the constructed ML regression. To optimize HEB, {we used the following optimization methods: \textit{surrogate-based optimization~(Sec.~\ref{sbo_bullet}), sequential least squares programming~\cite{slsqp}, particle swarm optimization~\cite{pso} and differential evolution~\cite{de} algorithms}}. The advantages of these methods are that they make no assumptions about the problem being optimized and can perform searches in  very large spaces of candidate solutions.

For our methodology, the goal is to maximize the cumulative fluid production by finding a set of optimal design parameters constrained by boundaries (see Sec.~\ref{boundaries}) for the specified parameters of the pilot well environment.

\subsection{Review of known optimization procedures}

\begin{figure}[h!]
\includegraphics[width=8cm]{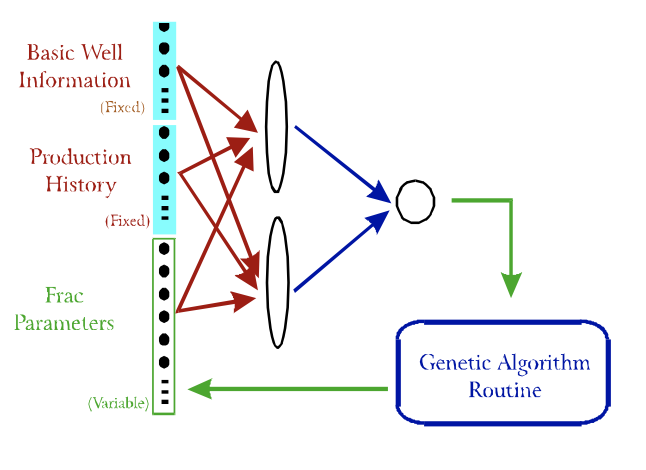}
\caption{A schematic diagram of the neuro-genetic approach}
\label{Neurogen}
\end{figure}

There are studies published in open literature on optimization of HF parameters based on some relatively moderate datasets with hundreds of wells, e.g., see a study on 570 wells in Clinton sand~\cite{mohaghegh1996modeling}. The first stage is a neural network which takes as input frac completions data and production history and predicts post-frac deliverability. The model  serves as a screening tool for excluding ``dog wells'', which cannot be considerably enhanced after a frac job. The second stage is based on a neural network, where the branch of the network, responsible for the fracture design parameters, is connected to the optimization algorithm to obtain  the optimized frac design for each well and the expected post frac deliverability (see Fig. \ref{Neurogen}).

The work~\cite{turbine} also utilizes a neuro-genetic algorithm  in the optimization pipeline instead of a computationally complex implicit numerical hydrodynamic model. In particular, the article presents a case study of neuro-genetic optimization using the example of a cross-flow microturbine. In another paper~\cite{chen2019} an ML regression analysis on more than 3500 wells was performed, the authors optimized the design by visual analysis with taking a pair of the most important parameters and selecting a region of the most optimal design.

Speaking about other industries, the article~\cite{SHI2019586} uses neural networks and a hybrid multisubgradient descent method with adaptive learning rate to solve multicriteria optimization of multiple tasks (generation of too large droplets and too low droplet speed) in the field of bioprinting. The proposed method can improve both printing accuracy and stability, and is useful for realizing precise cell arrays and complex biological functions.

Another application of machine learning and optimization is the design optimization of thin multilayer solar cells to maximize external quantum efficiency~\cite{kaya2019}. The authors utilize the concept of transfer learning which implies the use of experience gained in solving one problem to solve another, similar problem. The reason why the transfer learning is applied to the problem is because the problem involves the change of design specifications. Therefore, the transfer learning model acts as a function of surrogate optimization, which refits the surrogate more efficiently. In particular, the procedure improved the results by 2-3 times using only half of the training samples compared to the usual model.

\subsection{Surrogate-based optimization (SBO)}
\label{sbo_bullet}
Surrogate models (approximation models)~\cite{GTApprox2016} can be used with multi-dimensional input design spaces. As we know, the more parameters the surrogate model takes as input the more computational resources  (training time) it is required to construct the model. %During the optimization, we need to solve 2 problems: reduce the cost of surrogate model training and introduce multi-dimensional capabilities into the model. 

A particular example of Surrogate-based Optimization (SBO) or sequential model-based global optimization is Bayesian optimization. The algorithm utilizes a probabilistic surrogate function which approximates the expensive objective function and an acquisition function which, in its turn, allows to select a new candidate input design point within the domain. At the beginning, a surrogate function is built on a small number of samples from the original objective function, which provides a target value. Then, the algorithm maximizes the acquisition function and the surrogate is updated with a new input point and its actual output value. After repeating the process, an optimum can be achieved by taking the information about the samples from the past iterations. A typical choice of the surrogate is Gaussian processes and random forest, while typical acquisition functions are \textit{the Expected Improvement} (EI) and \textit{the Probability of Improvement} (PI).

We used the SBO algorithm implemented in the industrial software~\cite{p7}. The SBO methodology is based on Gaussian processes modeling technique \cite{ADoEGP,GPreg2016,MFGP2017,MFGP2015}. The particular numerical realization roots in the scientific works published in~\cite{ADoEGP}.

% удалил, так как нельзя использовать в стакнутой модели

% \subsection{Probability of improvement}
% For our methodology (Fig. \ref{Inverse_scheme}), the goal is to maximize the cumulative fluid production by finding a set of optimal design parameters constrained by boundaries (see Sec.~\ref{boundaries}) for the specified parameters of the environment. In addition to maximizing the actual target, given by the output of the constructed ML model, we propose to maximize a special criterion PI. The PI criterion is defined by an acquisition function:
% \begin{equation}
%     PI(x) \geq \Phi\left(\frac{\mu(x) - f(x+)}{\sigma(x)}\right),   
% \end{equation}
% where $f(x+)$ is the max predicted value already found during the optimization iterations, $\mu(x)$ is the current mean prediction of the current iteration, $\sigma(x)$ is the estimated standard deviation (uncertainty) obtained from \textit{CatBoost} of the current iteration, $\Phi$ is the cumulative probability function of a normal distribution. This allows us to be more certain about recommended optimal designs avoiding less certain ones.

% \begin{figure}[h!]
% \centering
% \includegraphics[width=7cm]{images/Inverse_problem_scheme.png}
% \caption{Inverse problem methodology scheme}
% \label{Inverse_scheme}
% \end{figure}

% \subsection{Gradient-free methods}
% \textcolor{red}{\bf Anton}
% \subsubsection{Target maximization}
% \subsubsection{Expected improvement}
% \begin{itemize}
%     \item Nelder-Mead
%     \item Trust-region
%     \item Simulated annealing
%     \item Genetic algorithm
%     \item Particle-swarm optimization
% \end{itemize}

\section{Results and Discussion}
\label{sec6}

During the research, we trained the model on wells from the field of interest only, thereby reducing the database from 6687 to 3308 fracturing operations.

{
Field tests have been carried out on 21 wells. 9 wells were horizontal, 7 --- vertical multilateral, operating on several layers simultaneously, the rest 5 wells were regular vertical ones. We were testing the accuracy of our prediction models, as well as HF design optimization overall pipeline. This pipeline includes:
\begin{enumerate}
    \item Obtaining design parameters optimization boundaries by similar wells search with DBSCAN clustering~(Sec.~\ref{boundaries});
    \item Imputing missing parameters with mean corresponding parameters values, calculated using top-10 similar wells, obtained within the pilot cluster via the euclidean distance search~(Sec.~\ref{offsetwells});
    \item Performing the design optimization, using the predictive model~(Sec.~\ref{sec3}) and optimization algorithms~(Sec.~\ref{design optimization})
\end{enumerate}
}

\subsection{Forward model prediction accuracy test}
\begin{figure}[h]
\centering
\includegraphics[width=12cm]{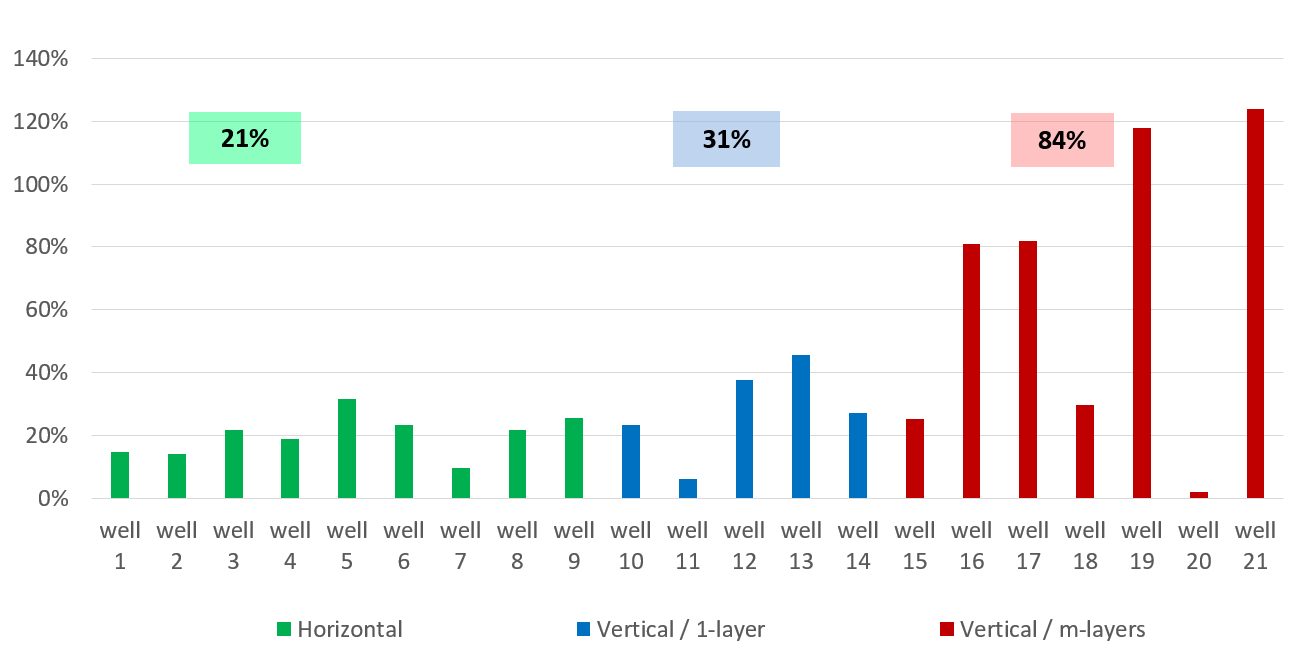}
\caption{{Real vs Predicted production on the real design}}
\label{Field_tests_perc}
\end{figure}

{
The production prediction errors are shown in~Fig.~\ref{Field_tests_perc}. Here we see a relatively low percentage error for horizontal wells, which can be explained by the higher production rates of such wells. What is important here is the high error for multilateral wells, which is most likely caused by data distortion for this type of wells. Currently, the data point for a multilateral well is represented as a multistage fracture treatment with the number of stages equal to the number of laterals with fractures. The disadvantages of this method are obvious when hydraulic fracturing is performed at different points in time. In addition, different reservoir parameters must be considered for each operational production formation.
}

{
Generally, the accuracy of our model ($MAPE=37.28\%$, $wMAPE=27.46\%$) in field tests is close to the hold-out set accuracy check ($MAPE=36.08\%$, $wMAPE=29.06\%$). The distribution of well types (vertical, horizontal) for field tests is close to that presented in the hold-out set.
}

\subsection{Design optimization test}
\label{design optimization}

Overall, we formulated four approaches to the problem of design optimization:
{
\begin{enumerate}
    \item \textit{SBO}: Surrogate-Based Optimization~(Sec.~\ref{sbo_bullet}),
    \item \textit{SLSQ}: Sequential Least Squares Programming,
    \item \textit{PSO}: Particle Swamp Optimization,
    \item \textit{DE}: Differential Evolution.
\end{enumerate}
}

We compared these methods with each other in terms of maximum produced fluid and physically-grounded recommendations for the design parameters for 3 well types: horizontal, vertical and vertical multilateral.

\begin{figure}[H]
\centering
\includegraphics[width=8cm]{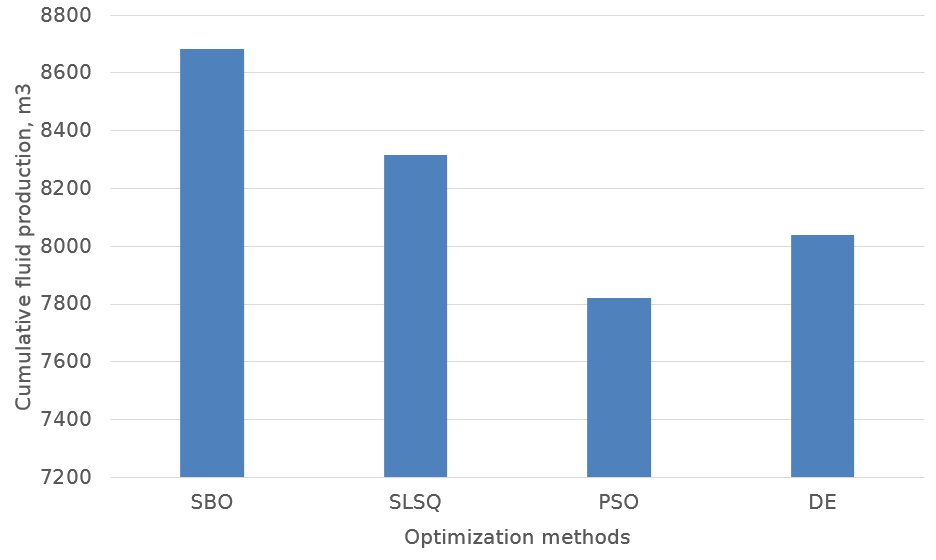}
\caption{{
Average optimized production for all wells by different optimization algorithms}}
\label{Target}
\end{figure}

\begin{figure}[h]
\centering
\includegraphics[width=13cm]{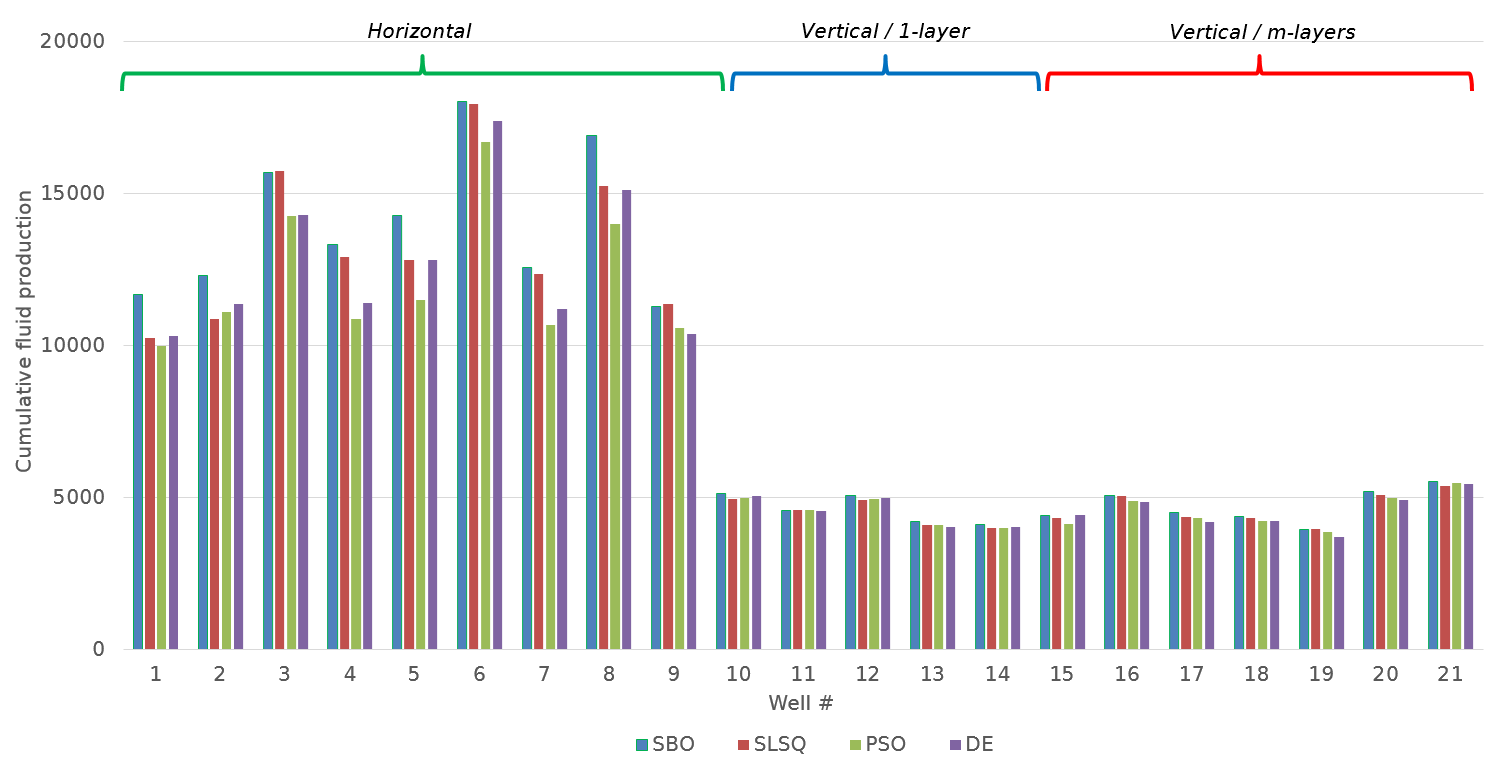}
\caption{{Optimized production comparison between methods}}
\label{prod_comp}
\end{figure}

{
Comparison of the efficiency of optimization algorithms in Figure~\ref{Target} shows the average cumulative fluid production across all pilot wells. A larger value means higher efficiency of an algorithm. Figure~\ref{prod_comp} shows the extended results of the optimization for each well individually.
}

\begin{figure}[H]
\centering
\includegraphics[width=8cm]{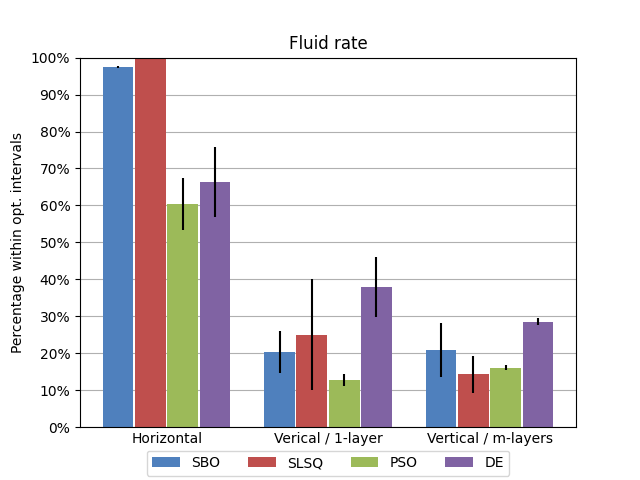}
\caption{{Average recommended fluid rate for the wells}}
\label{Fluid rate}
\end{figure}

\begin{figure}[H]
\centering
\includegraphics[width=8cm]{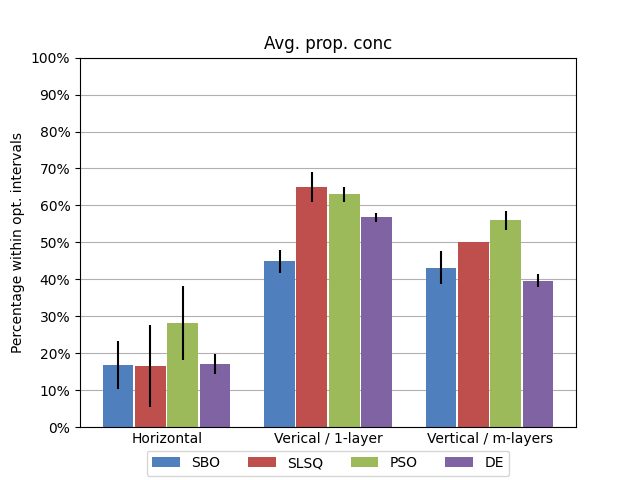}
\caption{{Average recommended mean proppant concentration for the wells}}
\label{Fluid sum}
\end{figure}

\begin{figure}[H]
\centering
\includegraphics[width=8cm]{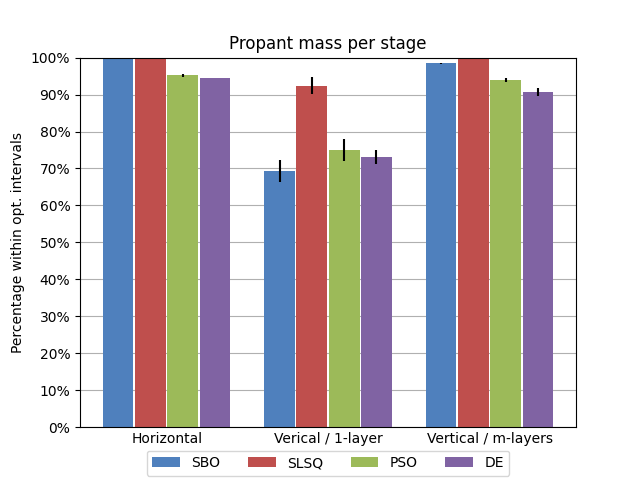}
\caption{{Average recommended proppant masses (per stage) for the wells}}
\label{Proppant sum}
\end{figure}

\begin{figure}[H]
\centering
\includegraphics[width=8cm]{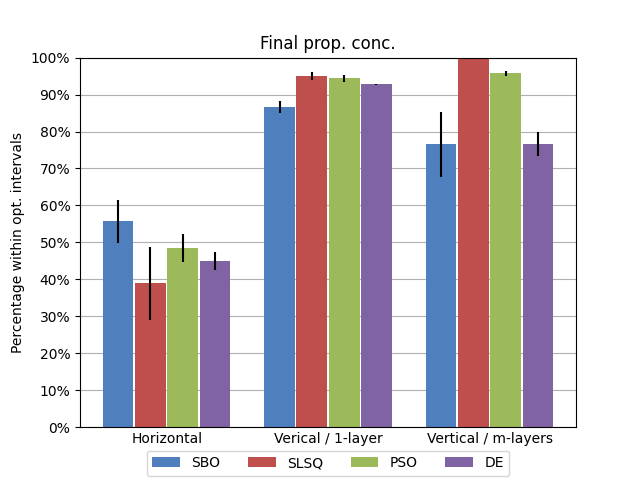}
\caption{{Average recommended final proppant concentration for the wells}}
\label{Proppant concentration}
\end{figure}

\begin{figure}[H]
\centering
\includegraphics[width=8cm]{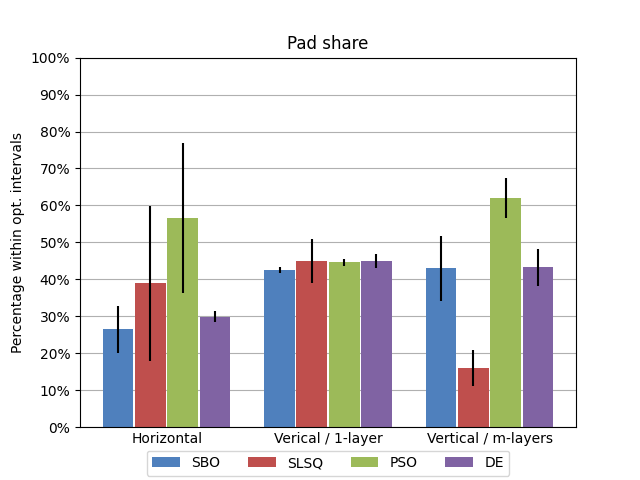}
\caption{{Average recommended pad share for the wells}}
\label{Pad share}
\end{figure}

\begin{figure}[H]
\centering
\includegraphics[width=8cm]{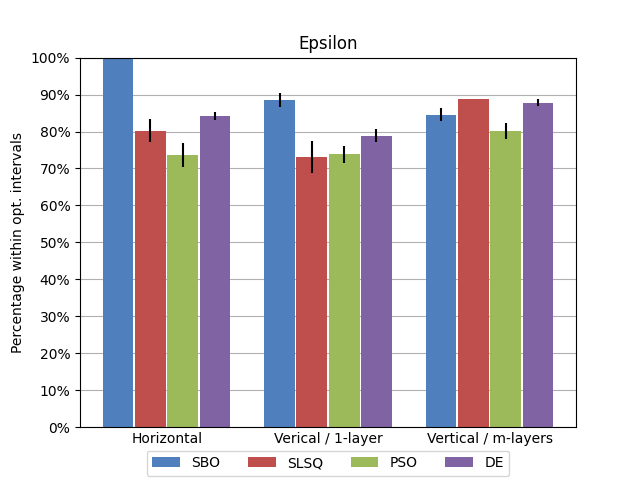}
\caption{{Average recommended calculated epsilon}}
\label{epsilon}
\end{figure}

{
The results of the optimization for design parameters are shown in Figures~\ref{Fluid rate}-\ref{epsilon}. Here we have indicated the recommended (optimum) values in percentages. 0\% is a lower and 100\% is an upper bound for each well, obtained by the offset wells search. Color bars represent the average value of the analyzed parameter for all wells in each of the three groups.}
% The average value of the recommended parameters is represented as a percentage within the boundaries, where 0\% and 100\% are the lower and upper optimization bounds for the particular well respectively.

{
All the methods were operating under similar conditions: boundaries, constraints, maximum allowable number of function evaluations (200). Under these conditions, SBO proved to be the most efficient algorithm in terms of maximizing the objective function~(Figures~\ref{Target}~and~\ref{prod_comp}).  
It is also possible to draw conclusions about general optimization trends. First of all, the most of the approaches for all well types maximized the target by increasing the proppant mass and reducing the average proppant concentration. It makes sense to increase the amount of pumping fluid and proppant to get maximum production while we have neither WOC nor GOC.
}

{
We can clearly see the recommendation trends difference between horizontal and both types of vertical wells even though vertical multilateral wells are represented in our database as multistage fracturing operations (like horizontal wells). Pad share here is the most interesting parameter, which does not follow any particular trend. The final proppant concentration is lower for horizontal wells, which can be explained by the high probability of ineffective treatments with high concentration values in these types of wells, as such fluids are difficult to pump through well sections with high wellbore curvature. The value of the fluid flow rate varies widely and is probably not important for the purpose of maximizing production. This can be proved by the low feature importance of this parameter, and may also be due to the fact that the values of this parameter are chosen properly in most of HF treatments presented in the database.
}

{Once we have the results of the optimization, we can carry out a kind of retrospective analysis to find out how we can change usual HF designs to improve its results: if the value is below 50\% -- the optimum value is less than usual one used during HF treatments and vice versa.}

% Robust recommendations (supported by experience of production stimulation engineers) are to maximize the fracturing fluid volume, along with the final proppant concentration and to set the pad share to about 30\% -- these design sets are very similar to those which are already realized in previous treatments on similar wells. Those parameter trends are recommended by the \textit{SBO} algorithm.

% However, in general, recommendations are to be reviewed by the experts, as there are a lot of subtleties in hydraulic fracturing design. For example, despite the recommendation to maximize the final proppant concentration, it is harder to perform in horizontal wells due to technical reasons. \textcolor{red}{These limits should be taken into consideration while obtaining the parameters optimization intervals via similar well search.}

\begin{figure}[h]
\centering
\includegraphics[width=12cm]{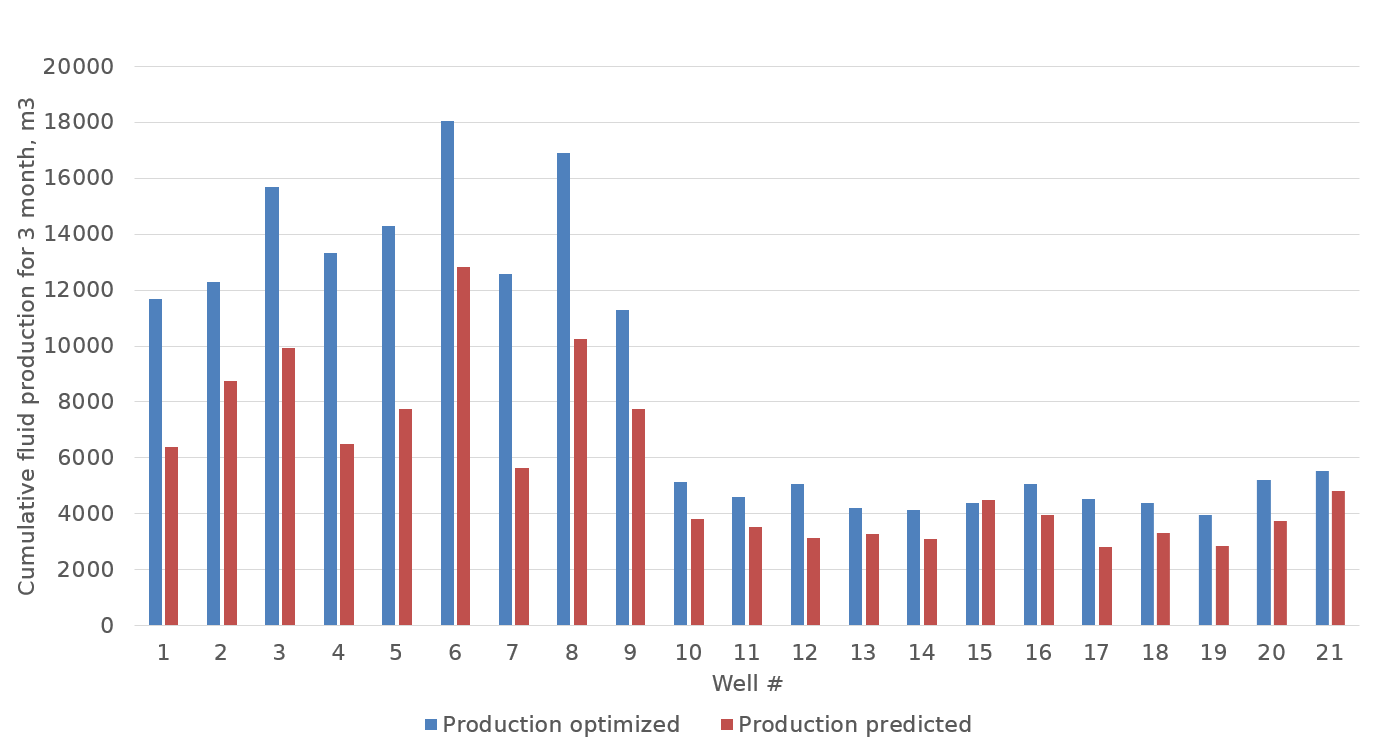}
\caption{{Production increase with optimal parameters set}}
\label{prod_increase}
\end{figure}

{
We tested the SBO method to find the optimal set of design parameters for each well from the pilot tests. In addition, we limited the proppant mass to a value from the actual fracture design (the actual values of the design parameters have been chosen by the contractor). We then predicted fluid production for the optimal parameters and for the actual ones. Optimal parameters showed a higher production rate, \textit{theoretically} increased by 38\%.~(Fig.~\ref{prod_increase}).
}
%--------------------------------------------------------------------------------------------

\section{Summary and Conclusions}
\label{sec7}

Since the first part of this work was published in~\cite{MLforHF2020}, both the digital database and the production forecast model have undergone several changes: the database was updated (now v. 7.9, compare with v. 5.5 in~\cite{MLforHF2020}) and new models were developed for testing. {The models for predicting cumulative fluid production now use a stacked approach with Ridge Regression and CatBoost trained on the residuals to get smooth dependencies of target values relative to design parameters. 
}

{
Field test were performed with 21 wells. The accuracy of our model in pilot tests  ($MAPE=37.28\%$, $wMAPE=27.46\%$) is close to the hold-out set accuracy check while model training ($MAPE=36.08\%$, $wMAPE=29.06\%$).
}

{
A method for similar wells search was developed. The method uses a clustering technique and the euclidean distance as a metric of similarity. The results help engineers to look back at previously conducted treatments. This method is also used to estimate the interval limits for the design parameters to be optimised. Lower and upper limits are $5^{th}$ and $95^{th}$ percentiles of the cluster parameter values respectively. One can reduce the size of this set of wells, leaving only top-N wells by the euclidean distance. This allows us to look for optimal values of the design parameters in a certain vicinity where the prediction model works well. Also, the top-N similar wells can be used for imputing missing parameters for the pilot well.
}

A number of optimization techniques were tested during the pilot testing: \textit{differential evolution, sequential least squares programming, particle swarm optimization and surrogate-based optimization}.

{
The \textit{SBO} approach on average maximized the 3-month fluid production better than other methods. General optimization trend of 21 pilot wells is to increase the amount of fracturing fluid and proppant mass. Trends for final, average proppant concentration and fluid rate are different for each type of well (horizontal, vertical and vertical multilateral). The calculated production from the fracturing with the optimal set of design parameters, compared to the treatments with an actual HF design, gives a theoretical target improvement of 38\%.
}

In future work, we plan the following extensions of the presented workflow: (i) extend the list of features to include presence of upper/lower water-bearing layers to be able to predict separately the production of pure oil and water, and extend the target to maximum total fluid and maximum pure oil (or minimum water cut); (ii) implement economics model to be able to work under the metrics in terms of Q/CAPEX; (iii) consider the influence of injection wells on production rates; (iv) extend the target further to be able to treat the problem as a multi-criteria optimization where the goal is to maximize the production and to simultaneously minimize the total proppant load. {
Furthermore, it can be very useful to combine synthetic data from fracture design simulations (i.e, using commercial simulators). With primary fracture design and reservoir properties data, we could simulate fracture propagation and obtain fracture length, width and height, which directly affect production. This should improve the predictive power of our models.
}

Finally, we plan to perform optimization for each well in the database to generalize the overall optimization trend.

\section*{Acknowledgements}
{The authors are grateful to the management of LLC ``Gazpromneft-STC'' for organizational and financial support of this work. The authors are particularly grateful to M.M.~Khasanov, A.A.~Pustovskikh, I.G.~Fayzullin and A.S.~Margarit for organizational support of this initiative. The help from P.K.~Kabanova and A.R.~Mukhametov in data gathering is greatefully appreciated. 

We are grateful to the management of the DATADVANCE company, namely, to S.M.~Morozov, and to the engineers of the company, A. Saratov and Yu.~Bogdanova, for the scientific support with Surrogate Based Optimization methods from the pSeven Core Suite, a product of DATADVANCE.

We would like to state it explicitly that the models presented in this work are solely based on the field data provided by JSC Gazpromneft and we are grateful for the permission to publish.}

\bibliographystyle{elsarticle-num}
\bibliography{MLfracturing}

\begin{thebibliography}{10}
\expandafter\ifx\csname url\endcsname\relax
  \def\url#1{\texttt{#1}}\fi
\expandafter\ifx\csname urlprefix\endcsname\relax\def\urlprefix{URL }\fi
\expandafter\ifx\csname href\endcsname\relax
  \def\href#1#2{#2} \def\path#1{#1}\fi

\bibitem{MLforHF2020}
A.~D. Morozov, D.~O. Popkov, V.~M. Duplyakov, R.~F. Mutalova, A.~A. Osiptsov,
  A.~L. Vainshtein, E.~V. Burnaev, E.~V. Shel, G.~V. Paderin, Data-driven model
  for hydraulic fracturing design optimization: Focus on building digital
  database and production forecast, Journal of Petroleum Science and
  Engineering 194 (2020) 107504.

\bibitem{RN2019refrac}
A.~Azbukhanov, I.~Kostrigin, K.~Bondarenko, M.~Semenova, I.~Sereda,
  D.~Yulmukhametov, et~al., Selection of wells for hydraulic fracturing based
  on mathematical modeling using machine learning methods (russian), Oil
  Industry Journal 2019~(11) (2019) 38--42.

\bibitem{xue2019shales}
L.~Xue, Y.~Liu, Y.~Xiong, Y.~Liu, X.~Cui, G.~Lei, A data-driven shale gas
  production forecasting method based on the multi-objective random forest
  regression, Journal of Petroleum Science and Engineering 196  107801.

\bibitem{yandex2020oil}
A.~Davtyan, A.~Rodin, I.~Muchnik, A.~Romashkin, Oil production forecast models
  based on sliding window regression, Journal of Petroleum Science and
  Engineering 195 (2020) 107916.

\bibitem{yao2021optimization}
J.~Yao, Z.~Li, L.~Liu, W.~Fan, M.~Zhang, K.~Zhang, Optimization of fracturing
  parameters by modified variable-length particle-swarm optimization in
  shale-gas reservoir, SPE Journal (2021) 1--18.

\bibitem{sobolPaper}
I.~Sobol, {Global sensitivity indices for nonlinear mathematical models and
  their Monte Carlo estimates}, {Mathematics and Computers in Simulation }\href
  {http://dx.doi.org/10.1016/S0378-4754(00)00270-6}
  {\path{doi:10.1016/S0378-4754(00)00270-6}}.

\bibitem{SHAP}
{Lundberg, Scott M., and Su-In Lee}, {A unified approach to interpreting model
  predictions}, Advances in Neural Information Processing Systems.

\bibitem{shapley1953value}
L.~S. Shapley, A value for n-person games, Contributions to the Theory of Games
  2~(28) (1953) 307--317.

\bibitem{song2016shapley}
E.~Song, B.~L. Nelson, J.~Staum, Shapley effects for global sensitivity
  analysis: Theory and computation, SIAM/ASA Journal on Uncertainty
  Quantification 4~(1) (2016) 1060--1083.

\bibitem{erofeev}
A.~Erofeev, D.~Orlov, D.~Perets, D.~Koroteev, {AI-Based Estimation of Hydraulic
  Fracturing Effect}, {SPE Journal }\href {http://dx.doi.org/10.2118/205479-PA}
  {\path{doi:10.2118/205479-PA}}.

\bibitem{wMAPE}
{Stephan Kolassa and Wolfgang Schütz}, {Advantages of the MAD/Mean Ratio over
  the MAPE}, Foresight: The International Journal of Applied Forecasting~(6)
  (2007) 40--43.

\bibitem{bellman2015applied}
R.~E. Bellman, S.~E. Dreyfus, Applied dynamic programming, Vol. 2050, Princeton
  university press, 2015.

\bibitem{slsqp}
Z.~Fu, G.~Liu, L.~Guo, {Sequential Quadratic Programming Method for Nonlinear
  Least Squares Estimation and Its Application}, Mathematical Problems in
  Engineering 2019 (2019) 1--8.
\newblock \href {http://dx.doi.org/10.1155/2019/3087949}
  {\path{doi:10.1155/2019/3087949}}.

\bibitem{pso}
M.~R. Bonyadi, Z.~Michalewicz, Particle swarm optimization for single objective
  continuous space problems: A review, Evolutionary Computation 25~(1) (2017)
  1--54.
\newblock \href {http://dx.doi.org/10.1162/EVCO_r_00180}
  {\path{doi:10.1162/EVCO_r_00180}}.

\bibitem{de}
R.~Storn, K.~Price, Differential evolution - a simple and efficient heuristic
  for global optimization over continuous spaces, Journal of Global
  Optimization 11 (1997) 341--359.
\newblock \href {http://dx.doi.org/10.1023/A:1008202821328}
  {\path{doi:10.1023/A:1008202821328}}.

\bibitem{mohaghegh1996modeling}
A.~S. Mohaghegh~S, Balan~B, A hybrid, neuro-genetic approach to hydraulic
  fracture treatment design and optimization.

\bibitem{turbine}
{E. T. Woldemariam, H. G. Lemu}, A machine learning based framework for model
  approximation followed by design optimization for expensive numerical
  simulation-based optimization problems, in: Proceedings of the Twenty-ninth
  International Ocean and Polar Engineering Conference www.isope.org Honolulu,
  Hawaii, USA, June 16-21, International Society of Offshore and Polar
  Engineers (ISOPE), 2019.

\bibitem{chen2019}
S.~Wang, S.~Chen, Insights to fracture stimulation design in unconventional
  reservoirs based on machine learning modeling, Journal of Petroleum Science
  and Engineering.

\bibitem{SHI2019586}
J.~Shi, J.~Song, B.~Song, W.~F. Lu, Multi-objective optimization design through
  machine learning for drop-on-demand bioprinting, Engineering 5~(3) (2019) 586
  -- 593.
\newblock \href {http://dx.doi.org/https://doi.org/10.1016/j.eng.2018.12.009}
  {\path{doi:https://doi.org/10.1016/j.eng.2018.12.009}}.

\bibitem{kaya2019}
M.~Kaya, S.~Hajimirza, Using a novel transfer learning method for designing
  thin film solar cells with enhanced quantum efficiencies, Nature.

\bibitem{GTApprox2016}
M.~Belyaev, E.~Burnaev, E.~Kapushev, M.~Panov, P.~Prikhodko, D.~Vetrov,
  D.~Yarotsky, Gtapprox: Surrogate modeling for industrial design, Advances in
  Engineering Software 102 (2016) 29 -- 39.

\bibitem{p7}
\href{https://www.datadvance.net/product/pseven/}{DATADVANCE© pSeven, Release
  6.20, User Manual}, 2021.
\newline\urlprefix\url{https://www.datadvance.net/product/pseven/}

\bibitem{ADoEGP}
E.~Burnaev, M.~Panov, Adaptive design of experiments based on gaussian
  processes, in: A.~Gammerman, V.~Vovk, H.~Papadopoulos (Eds.), Statistical
  Learning and Data Sciences, Springer International Publishing, Cham, 2015,
  pp. 116--125.

\bibitem{GPreg2016}
E.~Burnaev, M.~Panov, A.~Zaytsev, Regression on the basis of nonstationary
  gaussian processes with bayesian regularization, Journal of Communications
  Technology and Electronics 61~(6) (2016) 661--671.

\bibitem{MFGP2017}
A.~Zaytsev, E.~Burnaev, Large scale variable fidelity surrogate modeling,
  Annals of Mathematics and Artificial Intelligence 81~(1) (2017) 167--186.

\bibitem{MFGP2015}
E.~Burnaev, A.~Zaytsev, Surrogate modeling of multifidelity data for large
  samples, Journal of Communications Technology and Electronics 60~(12) (2015)
  1348--1355.

\end{thebibliography}
\onecolumn
\newpage
\centering
\section*{Appendix}

\end{document}